\documentclass{article}

\usepackage{setspace}

\usepackage[english]{babel}

\usepackage[letterpaper,top=2cm,bottom=2cm,left=3cm,right=3cm,marginparwidth=1.75cm]{geometry}
\usepackage{amsmath, amsthm, amssymb}
\usepackage{graphicx}
\usepackage{xcolor}
\usepackage[colorlinks=true, allcolors=blue]{hyperref}
\usepackage{subcaption}
\usepackage{mathtools}
\usepackage{wrapfig}
\usepackage{placeins}
\usepackage{float}
\usepackage{booktabs}
\usepackage{array}
\usepackage{multirow}

\usepackage[natbibapa]{apacite}

\graphicspath{{figures/}}

\usepackage{algorithm}
\usepackage{algpseudocode}

\newcommand{\emailfloris}{1}

\newif\ifshowrevisions
\showrevisionsfalse
\newcommand{\rev}[1]{\ifshowrevisions\textcolor{red}{#1}\else#1\fi}

\title{Selecting Hyperparameters for Tree-Boosting}
\author{
  Floris Jan Koster\footnotemark[1] \if1\emailfloris{\footnotemark[3] }\fi\\
  \and
  Fabio Sigrist\footnotemark[1] \footnotemark[2] 
}

\begin{document}
\date{}
\maketitle

\footnotetext[1]{Seminar for Statistics, ETH Zurich, Switzerland}
\footnotetext[2]{Corresponding author: fabio.sigrist@stat.math.ethz.ch}
\if1\emailfloris{
\footnotetext[3]{fkoster@ethz.ch }
}\fi

\begin{abstract}
Tree-boosting is a widely used machine learning technique for tabular data. However, its out-of-sample accuracy is critically dependent on multiple hyperparameters. In this article, we empirically compare several popular methods for hyperparameter optimization for tree-boosting including random grid search, the tree-structured Parzen estimator (TPE), Gaussian-process-based Bayesian optimization (GP-BO), Hyperband, the sequential model-based algorithm configuration (SMAC) method, and deterministic full grid search using $59$ regression and binary classification data sets. We find that the SMAC method clearly outperforms all the other considered methods on average, and it gives stable performance across a diverse collection of tabular data sets under a fixed tuning budget, which is relevant for users who cannot afford extensive manual trial-and-error tuning. We further observe that (i) a relatively large number of trials larger than $100$ is typically required for accurate tuning, (ii) using default values for hyperparameters or a deterministic search over a small grid often yields very inaccurate models, (iii) all considered hyperparameters can have a material effect on the accuracy of tree-boosting, i.e., there is no small set of hyperparameters that is more important than others, and (iv) choosing the number of boosting iterations using early stopping yields more accurate results compared to including it in the search space for regression tasks.
\end{abstract}

\noindent%

\section{Introduction}

Tree-boosting \citep{friedman2000additive, friedman2001greedy, buhlmann2007boosting, sigrist2018gradient} is a widely used machine learning technique that achieves state-of-the-art prediction accuracy on tabular data sets \citep{nielsen2016tree, shwartz2021tabular, januschowski2022forecasting, grinsztajn2022tree, mcelfresh2023neural}. However, the out-of-sample accuracy of tree-boosting depends critically on multiple hyperparameters, such as the number of trees and the learning rate. \rev{Despite a relatively large literature on hyperparameter optimization (HPO), selecting hyperparameters for tree-boosting remains a non-trivial practical problem: many applied studies still rely on default values, small ad-hoc grids, or a small number of trials. Moreover, previous comparisons often evaluated only a small number of data sets or HPO methods, or used different search spaces and tuning budgets.} In this article, we empirically compare several popular methods for selecting hyperparameters for tree-boosting using a large and diverse collection of regression and binary classification data sets. Specifically, we consider the following hyperparameter optimization methods: random grid search, the tree-structured Parzen estimator (TPE), Gaussian-process-based Bayesian optimization (GP-BO), Hyperband, the sequential model-based algorithm configuration (SMAC) method, a deterministic full grid search, and using default hyperparameters. 

We find that the SMAC method outperforms all other considered methods in terms of accuracy versus tuning budget. Moreover, we obtain the following findings. First, a relatively large number of trials (i.e., number of candidate hyperparameter sets) is required for accurate tuning, and using less than approximately $100$ trials often yields inaccurate models. Second, using default values for hyperparameters often yields very inaccurate predictions. Furthermore, we find that all considered hyperparameters have an effect on the accuracy of tree-boosting. This means that the sometimes observed practice of only tuning a small set of hyperparameters likely often results in inferior models. In addition, for regression tasks, we find that choosing the number of boosting iterations (i.e., the number of trees) using early stopping yields more accurate results compared to including the number of iterations in the search space of a hyperparameter selection method.

\subsection{Related literature}
\citet{putatunda2018hyperopt} compare the TPE method to random and deterministic full grid search for hyperparameter optimization for tree-boosting (XGBoost) on six classification data sets. They find that the TPE method achieves a superior accuracy–time trade-off compared to random and full grid search. \citet{motz2022benchmarking} compare hyperparameter optimization techniques on five industrial production data sets for various machine learning methods. They find that the SMAC and TPE methods perform best for tree-boosting. Recently, \citet{meaney2025comparison} compare multiple hyperparameter optimization methods for tree-boosting using a single healthcare data set with a relatively small number of features and a strong signal to noise ratio. They find that all methods yield similar results. In addition, there are public benchmark suites which include large collections of machine learning tasks for hyperparameter optimization comparisons such as \citet{eggensperger2021hpobench} and \citet{pfisterer22a}. Both \citet{eggensperger2021hpobench} and \citet{pfisterer22a} report that model-based search with resource allocation such as SMAC tends to outperform random or pure bandit baselines on tree-boosting scenarios. Several other articles have evaluated hyperparameter optimization (HPO) for general machine learning methods besides tree-boosting \citep[e.g.,][]{falkner2018bohb}. \citet{bergstraRandomSearch} find that random grid search outperforms deterministic grid search for neural networks. \rev{Our study complements these benchmarks by comparing default parameters, deterministic grids, random grids, TPE, GP-BO, Hyperband, and SMAC under a common search space and tuning budget on a large collection of regression and binary classification tabular data sets.}

\citet{van2018hyperparameter} analyze the importance of hyperparameters for various machine learning methods using functional ANOVA. They find for AdaBoost that most of the performance variation is explained mainly by the maximal depth of the decision tree and, to a lesser extent, the learning rate, whereas in our experiments, we find evidence that all considered hyperparameters appear to have a material effect on predictive accuracy. In contrast to \citet{van2018hyperparameter}, we consider a larger set of hyperparameters, gradient tree-boosting instead of AdaBoost, and we use a different importance-analysis methodology. \citet{probst2019tunability} analyze XGBoost from a different perspective based on ``tunability" of hyperparameters relative to default values and identify mainly \texttt{eta} and \texttt{booster} as the most tunable hyperparameters.

\section{Experimental settings}

\subsection{Hyperparameter selection methods, software and hardware used}\label{chp:Methods}

We consider the following methods for hyperparameter optimization: (i) deterministic full grid search, (ii) random grid search \citep{bergstraRandomSearch}, (iii) Gaussian-process-based Bayesian optimization (GP-BO) \citep{BO}, (iv) the tree-structured Parzen estimator (TPE) \citep{bergstra2011}, (v) Hyperband \citep{li2018hyperbandnovelbanditbasedapproach}, and (vi) sequential model-based algorithm configuration (SMAC) \citep{Hutter}. In addition, we compare this to using the default hyperparameters of \texttt{LightGBM} \citep{LightGBM} reported in Table \ref{table:default_pars}. 

Tree-boosting is done using the \texttt{GPBoost} Python package \citep{sigrist2021gpboost, sigrist2024} version 1.4.0 whose tree-boosting algorithm is the same as the one of \texttt{LightGBM} \citep{LightGBM}. For TPE and Hyperband, we use the \texttt{Optuna} Python package \citep{optuna_2019} version 3.5.0. For SMAC, we use the \texttt{SMAC3} Python package \citep{smac3} version 2.3.1, and for GP-BO, we use the \texttt{scikit-optimize} Python version 0.9.0. Unless stated otherwise, we use the default setting of all software packages. \rev{This choice reflects a practitioner-oriented comparison of commonly used HPO implementations rather than a fully customized version of each optimizer. It can introduce implementation-specific effects, for instance through different surrogate models or acquisition-function defaults, but it avoids tuning the optimizers themselves in a way that could favor one method.} For SMAC specifically, we use the \texttt{HPOFacade}, which is designed for standard hyperparameter optimization tasks and relies on a random forest surrogate model. The objective function is treated as deterministic, as the training procedure uses a fixed random seed. In particular, for all Bayesian optimization methods, we use the default acquisition functions in the above-mentioned software packages which are the expected improvement for TPE and SMAC and ``gp\_hedge" for GP-BO (each iteration randomly chooses among expected improvement, lower confidence bound, and probability of improvement). For SMAC, no specific multi-fidelity scheduling was enabled, and any early stopping was applied uniformly at the model-training level. Code to reproduce the experiments in this article can be found on \url{https://github.com/fl0risk/HPOTreeBoosting}.

All experiments were conducted using 32 dedicated physical CPU cores per job, with up to 32 GB of shared memory allocated per task. Computations were executed on high-performance computing nodes equipped with multi-core \texttt{AMD EPYC} processors.

\subsection{Data sets}\label{subsec:Dataset}
We use the same $59$ data sets as in \citet{grinsztajn2022tree} available on OpenML, out of which $36$ are regression and $23$ are classification tasks. \rev{This collection was chosen because it is a public and reproducible benchmark for tabular data sets and contains heterogeneous tasks that differ in sample size, number of features, signal strength, and response type.} We follow the pre-processing steps used in \citet{grinsztajn2022tree}. This includes dropping entries with missing values and removing categorical features with more than $20$ levels and numerical features with fewer than $10$ unique values. One-hot-encoding is used for categorical features. For the classification data sets, the response variable is binarized if there are multiple classes, by only including the two most prevalent classes. In contrast to \citet{grinsztajn2022tree}, we allow for larger sample sizes up to $n=100,000$. For data sets with more than $n=100,000$ samples, a random subsample of size $100,000$ is used.

\subsection{Train-test splits}\label{subsec:split}
We use $5$-fold cross-validation on every data set to compare the different methods. For every such $80/20$ train-test data split, we further split the training data into inner training and validation data sets using an $80/20$ split ratio, and the hyperparameters are chosen by learning on the inner training data and using a validation score on the validation data. As validation scores, the RMSE and accuracy are used for the regression and binary classification data sets, respectively. For a chosen set of hyperparameters, the models are then retrained on the entire training data sets. Most of the methods have a source of randomness such as randomly chosen initial values for the GP-BO, TPE, Hyperband, and SMAC methods and the random order in the random grid search. To analyze the impact of this, we repeat the hyperparameter searches $20$ times using different random number generator seeds. Specifically, for every data set and hyperparameter selection method, we use $20$ different random number generator seeds to generate $20$ random initial values for the adaptive methods and $20$ random orders for the random grid search. We then repeat the hyperparameter searches for each of the five $80/20$ train-test splits $20$ times using these random initial values and orders. 

\subsection{Hyperparameters and search spaces}\label{search_spaces}
We consider the following hyperparameters: the number of iterations (= number of trees), the learning rate, the number of leaves, the maximal depth, the minimal number of samples per leaf (`Min data in leaf'), the ${\ell_2}$ penalty on the leaf values (`Lambda ${\ell_2}$'), the maximal number of bins for the histogram-based splitting approach for continuous features (`Max bin'), and the bagging and feature sub-sampling fractions. 

For the TPE, GP-BO, Hyperband and SMAC methods, we use the hyperparameter search space given in Table \ref{table:base_search_space}. Note that ``$(l,u)$" denote intervals for continuous parameters, and ``$\{n_l,n_l+1,\dots,n_u-1,n_u\}$" denote sets of integers for discrete parameters. ``Max depth = -1" means no maximal tree-depth restriction, and $n$ denotes the sample size. For the random grid search, we use the hyperparameter grid shown in Table \ref{table:base_grid}. The hyperparameter grid for the deterministic full grid search is shown in Table \ref{tab:grid_gridsearch}, and the default hyperparameter values are reported in Table \ref{table:default_pars}. For the full grid search, default values are used for the hyperparameters not included in Table \ref{tab:grid_gridsearch}.

Both the maximal number of leaves and the maximal tree depth restrict the size of the trees. By default, we only include the maximal number of leaves in the hyperparameter search space and impose no limit on the maximal tree depth  (``Max depth = -1") since \texttt{LightGBM} uses a leaf-wise tree growth algorithm. However, we repeat the experiments by including the maximal depth in the search space and fixing the maximal number of leaves to a large number ($1024$) and also by jointly including the maximal number of leaves and the maximal tree depth in the hyperparameter search space. Furthermore, unless stated otherwise, we choose the number of boosting iterations using early stopping by monitoring a validation loss on the validation data sets. This means that the number of iterations is not explicitly contained in the search space of the hyperparameter selection methods, but for every combination of hyperparameters, the optimal number of iterations is determined using early stopping. To analyze the impact of this, we additionally perform the experiments by explicitly including the number of iterations in the search space instead of using early stopping. For this analysis, both the maximal number of leaves and the maximal tree depth are included in the search space, and for the random grid search, we use the set of candidate values for the number of iterations shown in Table \ref{table:base_grid}.

A total number of $135$ trials, which corresponds to the size of the deterministic full grid, is used in all experiments for all hyperparameter selection methods except for Hyperband. \rev{Thus, for random grid search, TPE, GP-BO, and SMAC, the main computational budget is directly comparable in terms of the number of fitted tree-boosting models. Since all methods tune the same underlying model on the same folds, differences in wall-clock time are mainly due to optimizer overhead and to the number of boosting iterations selected by early stopping. We did not record detailed wall-clock times or memory profiles for all runs; consequently, the empirical cost comparison in this article should be interpreted primarily through the common trial budget and the convergence curves of incumbent test scores, not as a complete runtime benchmark.} The {Hyperband} method has two main tuning parameters $R$ and $\eta$. We choose $R$ and $\eta$ such that the resulting maximal number of iterations is close to $135,000 = 135 \times 1,000$, which is the maximal number of boosting iterations without taking into account early stopping for the other methods. Specifically, we use $R = 2150$ and $\eta = 2.8$ which yields $128,816$ boosting iterations. \rev{This gives Hyperband a similar boosting-iteration budget to the other methods before early stopping is taken into account. As a coarse check for the chosen Hyperband resource parameters, the observed runtimes of random grid search and Hyperband were approximately equal across data sets (results not tabulated).}  
 \begin{table}[ht!]
    \centering
    \begin{tabular}{ll}
    \toprule 
        \textbf{Parameter} & \textbf{Search Space} \\
        \midrule
        Learning rate & $(0.001,\ 1)$ \\
        Num leaves & $ \{2,3,4,\dots 1023,1024\}$ \\
        Max depth & $ \{-1, 1, 2,3,\dots, 9,10\}$ \\
        Num iterations & $ \{1,2,\dots, 999, 1000\}$ \\
        Min data in leaf & $\{1,2,\dots, 999, 1000\}$ \\
        Lambda ${\ell_2}$  & $(0, 100)$ \\        
        Max bin & $\{255,256,257,\dots, \min\{10000,n\}\}$ \\
        Bagging fraction & $(0.5,\ 1)$ \\        
        Feature fraction & $(0.5,\ 1)$ \\
       \bottomrule
    \end{tabular}
    \caption{Hyperparameter search space for the {TPE}, {GP-BO}, {Hyperband}, and {SMAC} methods.}
    \label{table:base_search_space}
\end{table}

\begin{table}[ht!]
    \centering
     \begin{tabular}{ll}
        \toprule        
        \textbf{Parameter} & \textbf{Values} \\
        \midrule
        Learning rate & $\{0.001,\ 0.01,\ 0.1,\ 1\}$ \\ 
        Num leaves & $\{2,\ 4,\ 8,\ 32,\ 1024\}$ \\
        Max depth & $\{-1, 1,\ 2,\ 3,\ 5,\ 10\}$ \\
        Num iterations & $\{1, 2, 5, 10, 20, 50, 100, 200, 500, 1000\}$ \\
        Min data in leaf & $\{1,\ 10,\ 100,\ 1000\}$ \\
        Lambda ${\ell_2}$  & $\{0,\ 1,\ 10,\ 100\}$ \\        
        Max bin & $\{255,\ 500,\ 1000,\ \min\{10000,n\}\}$ \\
        Bagging fraction & $\{0.5,\ 0.75,\ 1\}$ \\        
        Feature fraction & $\{0.5,\ 0.75,\ 1\}$ \\
        \bottomrule
    \end{tabular}
    \caption{Hyperparameter grid for random grid search.}
    \label{table:base_grid}
\end{table}

\begin{table}[ht!]
    \centering
\begin{tabular}{ll}
        \toprule        
        \textbf{Parameter} & \textbf{Values} \\
        \midrule
        Learning rate & $\{0.01, 0.1, 1\}$ \\
        Min data in leaf & $\{10, 100, 1000\}$ \\
        Lambda $\ell_2$ & $\{0, 1, 10\}$ \\
        Num leaves & $\{2, 4, 8, 32, 1024\}$ \\
        \bottomrule
    \end{tabular}
    \caption{Hyperparameter grid for deterministic full grid search. Default values reported in Table \ref{table:default_pars} are used for the hyperparameters not included in this table.}
    \label{tab:grid_gridsearch}
\end{table}

\begin{table}[ht!]
    \centering
\begin{tabular}{ll}
        \toprule        
        \textbf{Parameter} & \textbf{Values} \\
        \midrule
        Learning rate & $0.1$ \\ 
        Num leaves & $31$ \\
        Max depth & $-1$ (=no limit) \\
        Num iterations & $100$ \\
        Min data in leaf & $20$ \\
        Lambda ${\ell_2}$  & $0$ \\  
        Lambda ${\ell_1}$  & $0$ \\   
        Max bin & $255$ \\
        Bagging fraction & $1$ \\        
        Feature fraction & $1$ \\
        \bottomrule
    \end{tabular}
    \caption{Default hyperparameters.}
    \label{table:default_pars}
\end{table}

\subsection{Evaluation scores and aggregation across data sets}\label{sec:metrics}
For evaluating the accuracy of the different methods, we use the following scores. For the regression tasks, we use the root mean squared error {(RMSE)} and test $R^2$ given by 
\begin{align*}
    \mathrm{RMSE} = \sqrt{\frac{1}{N} \sum_{i=1}^N (y_i - \hat{y}_i)^2}
\end{align*}
and
\begin{align*}
R^2 = 1- \frac{\mathrm{SS}_{\mathrm{res}}}{\mathrm{SS}_{\mathrm{tot}}},    
\end{align*}
respectively, where $\mathrm{SS}_{\mathrm{res}} = \sum_{i=1}^N (y_i - \hat{y}_i)^2$, $\mathrm{SS}_{\mathrm{tot}}= \sum_{i=1}^N (y_i - \overline{y})^2$, $y_i$ are the true values, $\hat{y}_i$ the predicted values, $\overline{y} = \frac{1}{N} \sum_{i=1}^N y_i$, and $N$ is the number of test samples. For the binary classification tasks, we use the {accuracy} and {log loss} given by
\begin{align*}
\mathrm{Accuracy} = \frac{\sum_{i=1}^N 1_{\{y_i = \hat{y}_i\}}}{\text{N}}
\end{align*}
and
\begin{align*}
    \text{Log loss} &= \frac{1}{N} \sum_{i = 1}^N -(y_i \log (p_i) + (1 - y_i) \log (1 - p_i)),
\end{align*}
respectively, where $p_i$ are the predicted probabilities for class $1$. As mentioned in Section \ref{subsec:split}, we use the RMSE and accuracy on the validation data sets for the regression and classification data sets, respectively, for choosing hyperparameters.


These scores are calculated on the hold-out test data sets after every trial $t$ using the currently best hyperparameters, which are determined only on the validation data sets for every method. Specifically, we calculate sequences of scores $s^t_{kmlj}$ for every data set $k=1,\dots,59$, hyperparameter selection method $m=1,\dots,7$, random number generator seeds $l=1,\dots,L_m$, train-test split $j=1,\dots,5$, and trial number $t=1,\dots,T_m$. For all random methods, $L_m=20$, and for the deterministic methods (default values and deterministic grid), $L_m=1$. For all hyperparameter optimization methods except Hyperband and using default values, the number of trials is $T_m=135$. For Hyperband, we calculate the scores $s^t_{kmlj}$ for $T_m=9$ rungs as follows. Hyperband runs $s_{\max}+1$ brackets consisting of multiple randomly sampled hyperparameter configurations. For each bracket, the algorithm allocates resources (number of boosting iterations) across configurations, and it performs successive halving: after evaluating configurations at a given budget, the poorest performers are discarded and the budget is increased for the remaining configurations, typically until a single configuration remains. After each successive-halving rung, we record the overall incumbent configuration (best validation performance among all configurations evaluated so far across all brackets) and use it to compute the corresponding test score $s^t_{kmlj}$. In our case, this results in $T_m=9$ recorded rungs.

For aggregating the scores $s^t_{kmlj}$ across the different data sets, we normalize them for better comparability across data sets as described in the following. First, we calculate averages over the five different folds:
$$s^t_{kml} = \frac{1}{5}\sum_{j=1}^5s^t_{kmlj}.$$
We then follow \citet{grinsztajn2022tree} and use the {average distance to the minimum} (ADTM) normalization \citep{ADTM}. Specifically, for the scores where lower values are better (RMSE and log loss), we use the ADTM normalization
\begin{align*}
    \tilde s^t_{kml} = \frac{ s^t_{kml} - \text{min}_k}{ Q^{0.9}_k - \text{min}_k},
\end{align*}
where ``$\text{min}_k$" and ``$Q^{0.9}_k$" denote the minimum and $90\%$ quantile of all scores $\{s^t_{kml}: m=1,\dots,7, l=1,\dots,20, t=1,\dots,T_m\}$ for task $k=1,\dots,59$. We use the $90\%$ quantile instead of the maximum since a few scores are very large in the first few trials of some methods. Using the maximum would distort the normalization in the sense that most normalized scores were close together and small. Analogously, for the scores where higher values are better ($R^2$ and accuracy), we use the ADTM normalization
\begin{align*}
    \tilde s^t_{kml} = \frac{ s^t_{kml} - Q^{0.1}_k }{ \text{max}_k - Q^{0.1}_k},
\end{align*}
where ``$\text{max}_k$" and ``$Q^{0.1}_k$" denote the maximum and $10\%$ quantile of all scores $\{s^t_{kml}: m=1,\dots,7, l=1,\dots,L_m, t=1,\dots,T_m\}$ for task $k$. Similarly as for the lower-better scores, we use the $10\%$ quantile instead of the minimum since a few scores are very small in the first few trials of some methods. Note that the results presented in this article are not sensitive to the specific choice of the $10\%$ and $90\%$ quantiles, and other quantiles yield qualitatively very similar results (results not shown). Finally, we calculate averages across the data sets and random number generator seeds
$$
\tilde s^t_{m} = \frac{1}{K}\frac{1}{L_m}\sum_{k=1}^{K}\sum_{l=1}^{L_m}\tilde s^t_{kml},
$$
where $K=36$ and $K=23$ for the regression and classification data sets, respectively,

In addition to the ADTM-normalized scores, we use ranks and relative differences to the best score for comparison across data sets. Specifically, ranks $r^t_{km}$ and relative differences to the best score $\delta^t_{km}$ are calculated using the sets $\{s^t_{km}: m=1,\dots,7\}$ for every iteration $t$ and data set $k$, where $s^t_{km} = \frac{1}{L_m}\sum_{l=1}^{L_m}s^t_{kml}$ are average scores over the different random number generator seeds. These ranks and relative differences are then averaged across data sets to give $r^t_{m}=\frac{1}{K}\sum_{k=1}^{K}r^t_{km}$ and $\delta^t_{m}=\frac{1}{K}\sum_{k=1}^{K}\delta^t_{km}$. However, relative differences have the disadvantage of being sensitive to the scale. For instance, two methods with accuracies of $0.1$ and $0.2$ have a large relative difference. However, when equivalently using error rates (given by $0.9$ and $0.8$ in this example) instead of accuracies, the relative difference becomes small. Ranks, on the other hand, have the disadvantage that they potentially neglect useful information in the sense that ranked scores are different even if the differences are tiny and practically negligible. 

\section{Results} \label{sec:comparison}
\begin{figure}[ht!]
    \centering
    \includegraphics[width=0.8\textwidth]{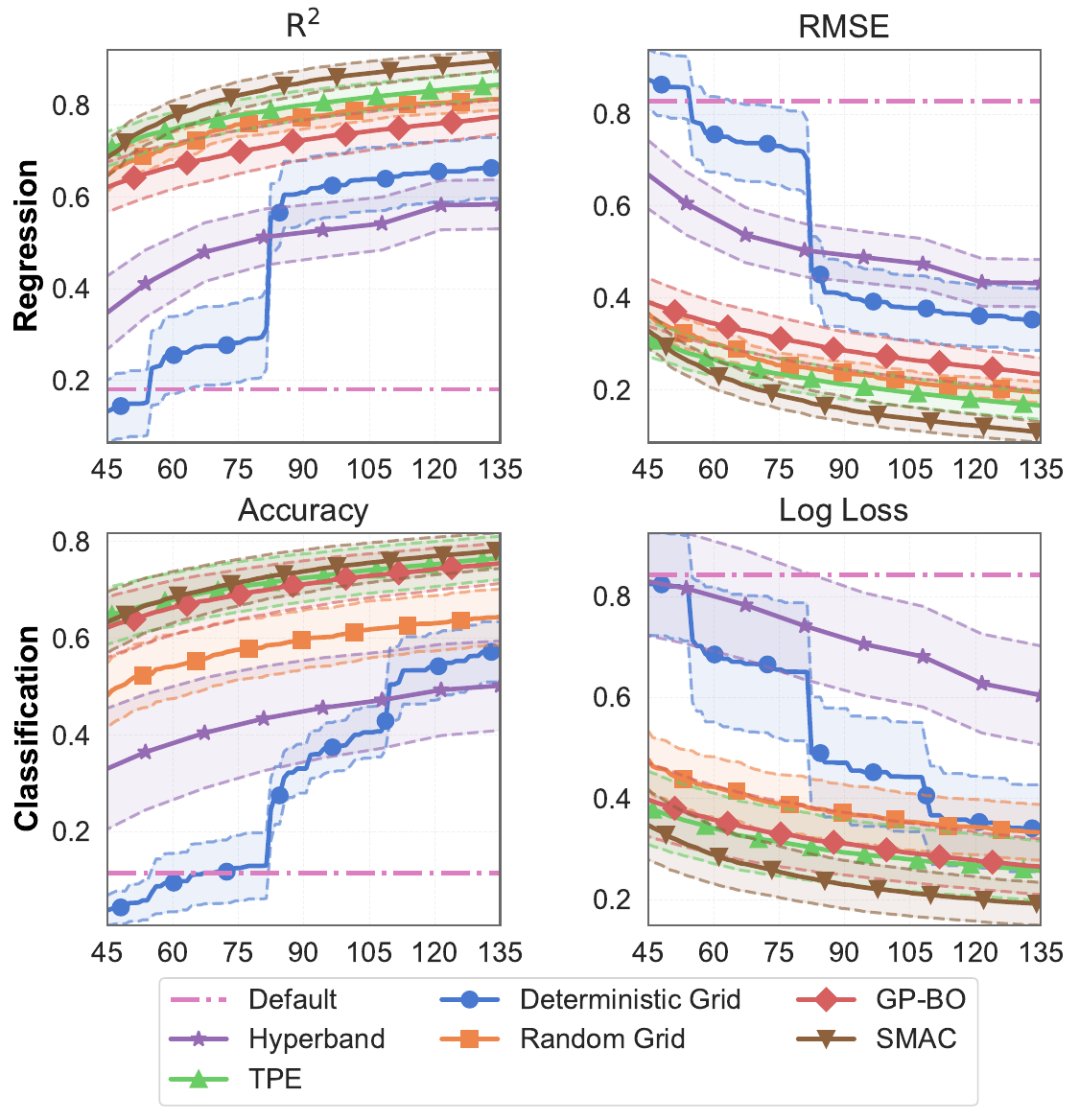}
    \caption{Normalized test accuracy scores as a function of the number of tuning trials for different HPO methods. \rev{The confidence intervals represent uncertainty across data sets after averaging over random initial values and random orders; they do not represent the randomness of repeated optimizer runs, which is shown separately in Figure \ref{fig:comparison_num_leaves_seed} in Appendix \ref{appendix}.}}
    \label{fig:comparison_num_leaves_task}
\end{figure}

Table \ref{tab:table_compare_methods_num_leaves} reports the average ADTM-normalized scores, relative differences to the best score, and ranks aggregated across all data sets after the maximal number of trials $T_m$. Specifically, the table contains the average normalized scores $\tilde s^T_{m}$, relative differences $\delta^T_{m}$, and ranks $r^T_{m}$ for the final best hyperparameters obtained at the end of the optimization for every method; see Section \ref{sec:metrics} for more details. Moreover, Figure \ref{fig:comparison_num_leaves_task} shows the sequences $\tilde s^t_{m}$ of average normalized incumbent scores as a function of the number of trials $t$. Note that we exclude the first $44$ trials as some methods yield very inaccurate results which would impair the visibility of the plots. We add $95$\% confidence intervals obtained as $\tilde s^t_{m}\pm 1.96 \text{SE}^t_m$, where $\text{SE}^t_m$ are standard errors representing uncertainty across data sets obtained by first averaging the normalized scores $\tilde s^t_{kml}$ over the different random number generator seeds, $\tilde s^t_{km} = \frac{1}{L_m}\sum_{l=1}^{L_m}\tilde s^t_{kml}$, and then using this to calculate standard errors across the data sets $k$. That is, the confidence intervals represent uncertainty across data sets but not uncertainty due to randomness in the hyperparameter selection methods. Below, we also analyze the latter uncertainty. In Figure \ref{fig:relative_comparison_methods_num_leaves_not_normalized_None} in Appendix \ref{appendix}, we additionally report the average relative differences $\delta^t_{m}$ as a function of the number of trials $t$. Recall that for the Hyperband method, there are not $T_m=135$ trials but $T_m=9$ rungs as explained in Section \ref{sec:metrics}. For better visual comparability, we uniformly place the $T_m=9$ rungs on the x-axis and linearly interpolate the test scores obtained by Hyperband in all figures that show the accuracy measures as a function of the number of trials $t$ such as Figures \ref{fig:comparison_num_leaves_task} and \ref{fig:relative_comparison_methods_num_leaves_not_normalized_None}.

\begin{table}[ht!]
\begin{tabular}{llccccccc}
\toprule
 && Default & Deterministic & GP-BO & Hyperband & Random Grid & SMAC & TPE \\
\midrule
\multirow{4}{*}{\rotatebox[origin=c]{90}{norm}} 
&$\mathrm{R}^2$  & 0.181 & 0.664 & 0.775 & 0.584 & 0.814 & \textbf{0.898} & 0.845 \\
&RMSE  & 0.829 & 0.353 & 0.234 & 0.432 & 0.195 & \textbf{0.108} & 0.166 \\
&Accuracy  & 0.115 & 0.572 & 0.755 & 0.502 & 0.645 & \textbf{0.782} & 0.766 \\
&Log Loss  & 0.843 & 0.341 & 0.265 & 0.604 & 0.333 & \textbf{0.191} & 0.257 \\
\midrule
\multirow{4}{*}{\rotatebox[origin=c]{90}{Rel. diff.}} 
&$\mathrm{R}^2$  & 0.046 & 0.013 & 0.006 & 0.017 & 0.004 & \textbf{0.001} & 0.003 \\
&RMSE  & 0.154 & 0.062 & 0.014 & 0.059 & 0.013 & \textbf{0.004} & 0.022 \\
&Accuracy  & 0.029 & 0.007 & 0.001 & 0.009 & 0.004 & \textbf{0.000} & 0.002 \\
&Log Loss  & 0.222 & 0.051 & 0.019 & 0.097 & 0.037 & \textbf{0.005} & 0.023 \\
\midrule
\multirow{4}{*}{\rotatebox[origin=c]{90}{Rank}} 
&$\mathrm{R}^2$  & 6.889 & 4.444 & 3.722 & 5.778 & 3.167 & \textbf{1.417} & 2.583 \\
&RMSE  & 6.889 & 4.472 & 3.694 & 5.778 & 3.167 & \textbf{1.417} & 2.583 \\
&Accuracy  & 6.783 & 4.870 & 2.304 & 5.652 & 4.087 & \textbf{1.870} & 2.435 \\
&Log Loss  & 6.652 & 3.696 & 2.957 & 6.043 & 3.826 & \textbf{1.652} & 3.174 \\
\bottomrule
\end{tabular}
\caption{Average normalized scores, relative differences, and ranks. }
\label{tab:table_compare_methods_num_leaves}
\end{table}

We find that {SMAC} clearly outperforms all other methods in terms of all metrics for both regression and binary classification tasks. \rev{A plausible explanation is that SMAC's random-forest surrogate can handle mixed discrete and continuous search spaces and non-smooth validation losses comparatively well, while still using previous evaluations to guide later trials.} Overall, the TPE method gives the second most accurate results. For the regression data sets, random grid search is almost equally accurate as the TPE method, followed by Gaussian-process-based Bayesian optimization (GP-BO). For the classification tasks, GP-BO yields essentially equal accuracy as the TPE method, and random grid search is less accurate. A deterministic full grid search and Hyperband give considerably worse results. \rev{One possible explanation for the weaker performance of Hyperband is that its successive-halving mechanism may be less effective when low-resource evaluations are only weakly informative about high-resource performance. In tree-boosting with early stopping and noisy validation losses, early low-budget evaluations might discard some configurations that would have become competitive after more boosting iterations.} Moreover, using default values for the hyperparameters results in very inaccurate predictions. Finally, we also observe that using a low number of trials, say below $100$, in the hyperparameter selection method yields worse results for all methods.

To assess the variability due to randomness in the methods, we additionally show in Figure \ref{fig:comparison_num_leaves_seed} in Appendix \ref{appendix} ``seed randomness" confidence intervals obtained by first averaging over the different data sets, $\frac{1}{K}\sum_{k=1}^{K}\tilde s^t_{kml}$, and then calculating standard errors based on these averages. Despite the relatively small number of different random initial values and orders ($20$), the confidence intervals are very small, and we conclude that the uncertainty in our results due to randomness in the methods is almost negligible.

In Figures \ref{fig:comparison_by_task_num_leaves_R2_1}, \ref{fig:comparison_by_task_num_leaves_R2_2}, \ref{fig:comparison_by_task_num_leaves_RMSE_1}, \ref{fig:comparison_by_task_num_leaves_RMSE_2}, \ref{fig:comparison_by_task_num_leaves_acc}, and \ref{fig:comparison_by_task_num_leaves_log_loss} in Appendix \ref{appendix} we additionally report the $R^2$, RMSE, accuracy, and log-loss as a function of the number of trials for every data set separately. No normalization is applied here as the results are not aggregated over the different data sets. In line with the results reported above, SMAC often yields the most accurate results for most data sets. However, there is some variability across the different data sets, and no method is universally best for all data sets.

The results discussed so far are obtained by including the maximal number of leaves in the hyperparameter search space without directly limiting the maximal tree depth and by choosing the number of boosting iterations using early stopping as described in Section \ref{search_spaces}. In Figure \ref{fig:comparison_tuning_grid_regr} in Appendix \ref{appendix}, we report additional results obtained by (i) including the maximal tree depth in the search space and fixing the maximal number of leaves to a large number (``Max Depth"), (ii) jointly including the maximal number of leaves and the maximal tree depth in the hyperparameter search space (``Joint"), and (iii) additionally including the number of iterations in the search space instead of using early stopping (``Num Iter"). For comparison, the figure also reports the results when including the maximal number of leaves in the hyperparameter search space without imposing a maximal tree depth limit (``Num Leaves"). We first observe that SMAC yields the best results irrespective of which hyperparameter search space option is used. Overall, including the maximal number of leaves in the hyperparameter search space without imposing an explicit limit on the maximal tree depth yields the best results. Including the number of iterations in the search space of the methods instead of using early stopping clearly yields worse results for the regression data sets. On the other hand, including the number of iterations in the search space gives the best results for the classification tasks for the random grid search, TPE, and SMAC methods.


\section{The importance of individual hyperparameters}
In the following, we try to understand whether some hyperparameters are more important than others for the prediction accuracy of tree-boosting. This is motivated by the observation that some empirical studies in various applied fields only tune a few hyperparameters while fixing others at default values. For this, we create a ``meta" data set consisting of all hyperparameter combinations and validation losses in all trials done for all data sets and methods in the above reported experiments when including both the maximal number of leaves and the maximal tree depth in the search space and using early stopping for the number of boosting iterations. Note that the hyperparameters are the predictor variables and the validation loss is the response variable in this data set. We then analyze how the validation loss depends on the hyperparameters using a tree-boosting regression model. To account for heterogeneity across data sets, we fit a mixed-effects tree-boosting model \citep{sigrist2024} with data-set-specific grouped random effects to this meta data set. \rev{Concretely, we include a data-set-level grouped random intercept, so that differences in the baseline validation loss across tasks are modeled separately from the fixed effects of the hyperparameters.} We apply the SMAC method to find the hyperparameters using an $80/20$ train-test split and the ``maximal number of leaves" search space option described in Section \ref{search_spaces} and early stopping for choosing the number of boosting iterations in this meta analysis. To quantify the importance of the different hyperparameters, we use SHAP values \citep{lundberg2017unified} \rev{computed from the tree-boosting model trained on this meta data set, where the validation loss is the response variable and the trial hyperparameters are predictor variables. Because the mixed-effects model uses data-set-level random intercepts, the resulting SHAP analysis should be interpreted as summarizing average hyperparameter effects across tasks rather than task-specific importance profiles.} This approach is similar in spirit to the functional ANOVA method of \citet{hutter2014efficient}, but uses SHAP values rather than a variance decomposition, thereby providing local additive attributions that can be aggregated to quantify global hyperparameter importance. \rev{A potential limitation of this meta-analysis is that the pooled trial data are not sampled uniformly from the full hyperparameter search space: adaptive optimizers evaluate configurations depending on previous validation results. Consequently, the SHAP values describe hyperparameter importance for the configurations explored in our benchmark, but they should not be interpreted as importance measures for an idealized uniform exploration of the entire search space.}

\begin{figure}[ht!]
    \centering
    \includegraphics[width=\linewidth]{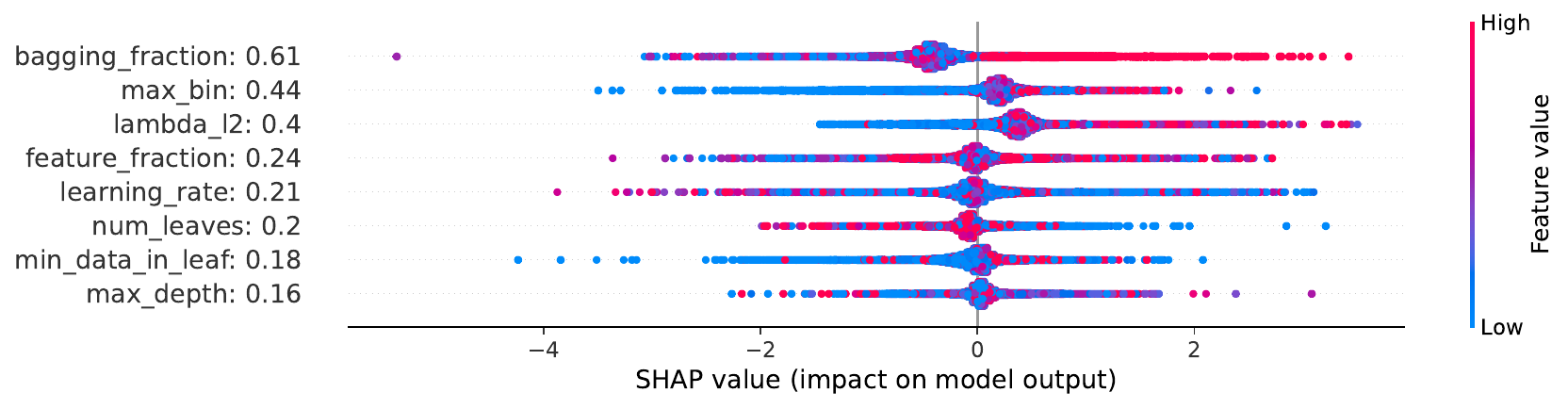}
    \includegraphics[width=\linewidth]{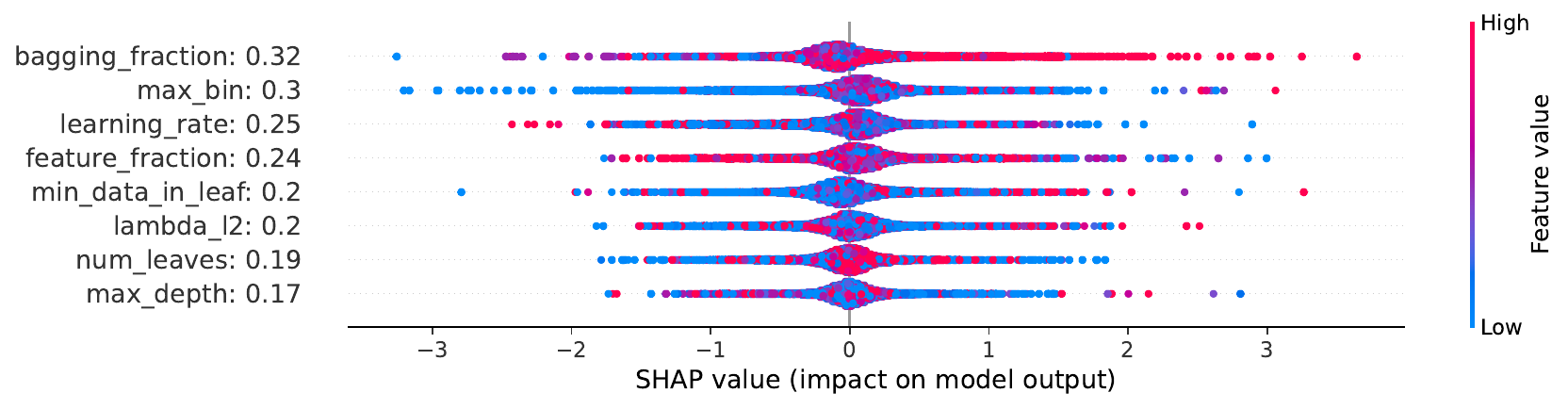}
    \caption{SHAP summary plot illustrating the impact of hyperparameters for regression (top) and classification (bottom) tasks.}
    \label{fig:shap_summary}
\end{figure}

In Figure \ref{fig:shap_summary}, we report SHAP values for the tree-boosting model trained on the meta data set described above. Separate SHAP values are reported for the regression and binary classification tasks. We find that all hyperparameters have relatively large average absolute SHAP values and the differences in the SHAP values are \rev{relatively} small across the hyperparameters. There is nevertheless some visible variation in average absolute SHAP values, so our conclusion is not that all hyperparameters have equal importance. For the classification meta-model, for instance, the largest average absolute SHAP value is about $1.85$ times the smallest. Rather, the plot shows no dominant subset whose importance is so large that the remaining parameters can safely be fixed at default values. This suggests that all considered hyperparameters can have a large effect on the accuracy of a tree-boosting model. 

\section{Conclusion}
In this article, we benchmarked several widely used hyperparameter optimization methods for tree-boosting on 59 OpenML regression and binary classification tasks under a common evaluation protocol and a comparable tuning budget. Overall, the results show a clear ranking: SMAC consistently achieves the best predictive performance across all considered metrics, with TPE typically being the second strongest approach, while GP-based Bayesian optimization and random grid search form a competitive middle tier depending on the task type. In contrast, deterministic grid search and Hyperband are markedly less reliable, and default hyperparameters often lead to substantially inferior accuracy. 

Beyond this headline comparison, several practical lessons emerge. First, accurate tuning generally requires a non-trivial number of trials: performance often continues to improve up to (and beyond) roughly 100 trials for methods such as SMAC, TPE, GP-based Bayesian optimization, and random grid search. This means that small trial budgets can materially distort conclusions about both models and tuning methods. Second, our analysis of hyperparameter importance suggests that there is no single “small” subset of hyperparameters that can be tuned while safely leaving others at defaults—all investigated parameters (learning rate, tree size/regularization controls, histogram binning, and subsampling) meaningfully affect performance, implying that partial tuning strategies are frequently suboptimal. Third, regarding the number of boosting iterations, we find that selecting it via early stopping is generally preferable to treating it as a standard search parameter for regression tasks, whereas for classification tasks including the iteration count in the search space can be competitive for some hyperparameter optimization methods.

Taken together, these findings support a simple recommendation for practitioners who can afford moderate tuning effort: \rev{use SMAC (or a closely related model-based method) when a stable, general-purpose optimizer is needed, allocate enough trials for the incumbent performance to stabilize, avoid relying on defaults, and prefer early stopping for determining the number of boosting rounds in regression tasks.} At the same time, the best method is not universal at the individual data-set level, so the observed average advantages should be interpreted as guidance rather than a guarantee.

There are also limitations that point to future work. Our experiments focus on one tree-boosting implementation (LightGBM), and we primarily assess predictive performance rather than full cost–benefit trade-offs (e.g., wall-clock time under varying degrees of parallelism, robustness under strict compute limits, or tuning under alternative objectives such as calibration or fairness). Future studies could extend these comparisons to additional boosting libraries, investigate principled ways to allocate budgets across folds and seeds, explore hybrid approaches that combine multi-fidelity resource allocation with strong model-based search and meta-learning/warm-starting across related tasks, \rev{and record detailed runtime and resource-consumption profiles.}

\section*{Acknowledgments}
This research was partially supported by the Swiss Innovation Agency - Innosuisse (grant number `57667.1 IP-ICT'). 

\bibliographystyle{apacite}
\bibliography{myReferences}

\appendix
\section*{Appendix}
\section{Additional results}\label{appendix}

\begin{figure}[ht!]
    \centering
    \includegraphics[width=0.8\linewidth]{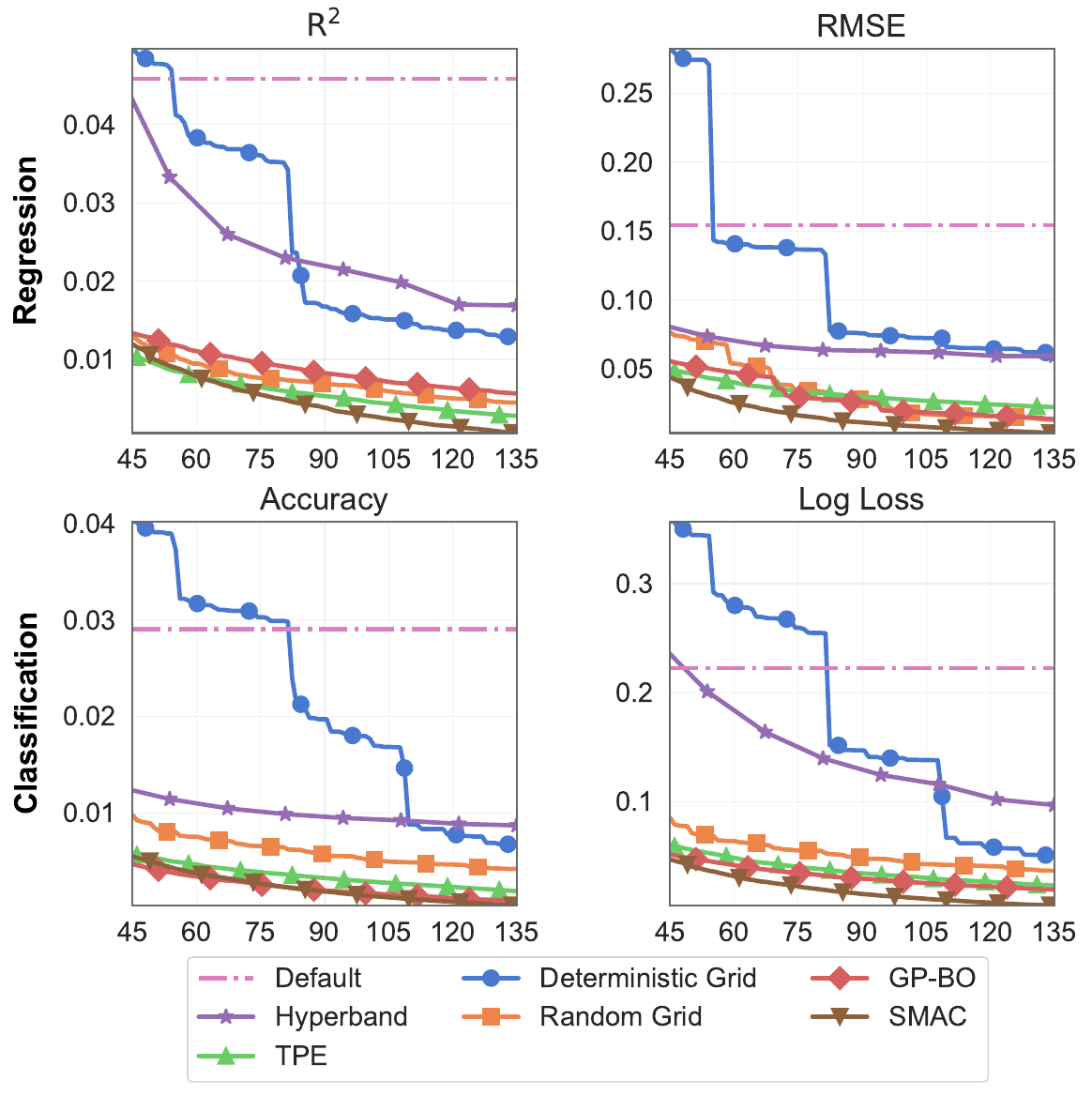}
    \caption{Average relative differences to the best score as a function of the number of trials.}
    \label{fig:relative_comparison_methods_num_leaves_not_normalized_None}
\end{figure}


\begin{figure}[ht!]
    \centering
    \includegraphics[width=0.8\linewidth]{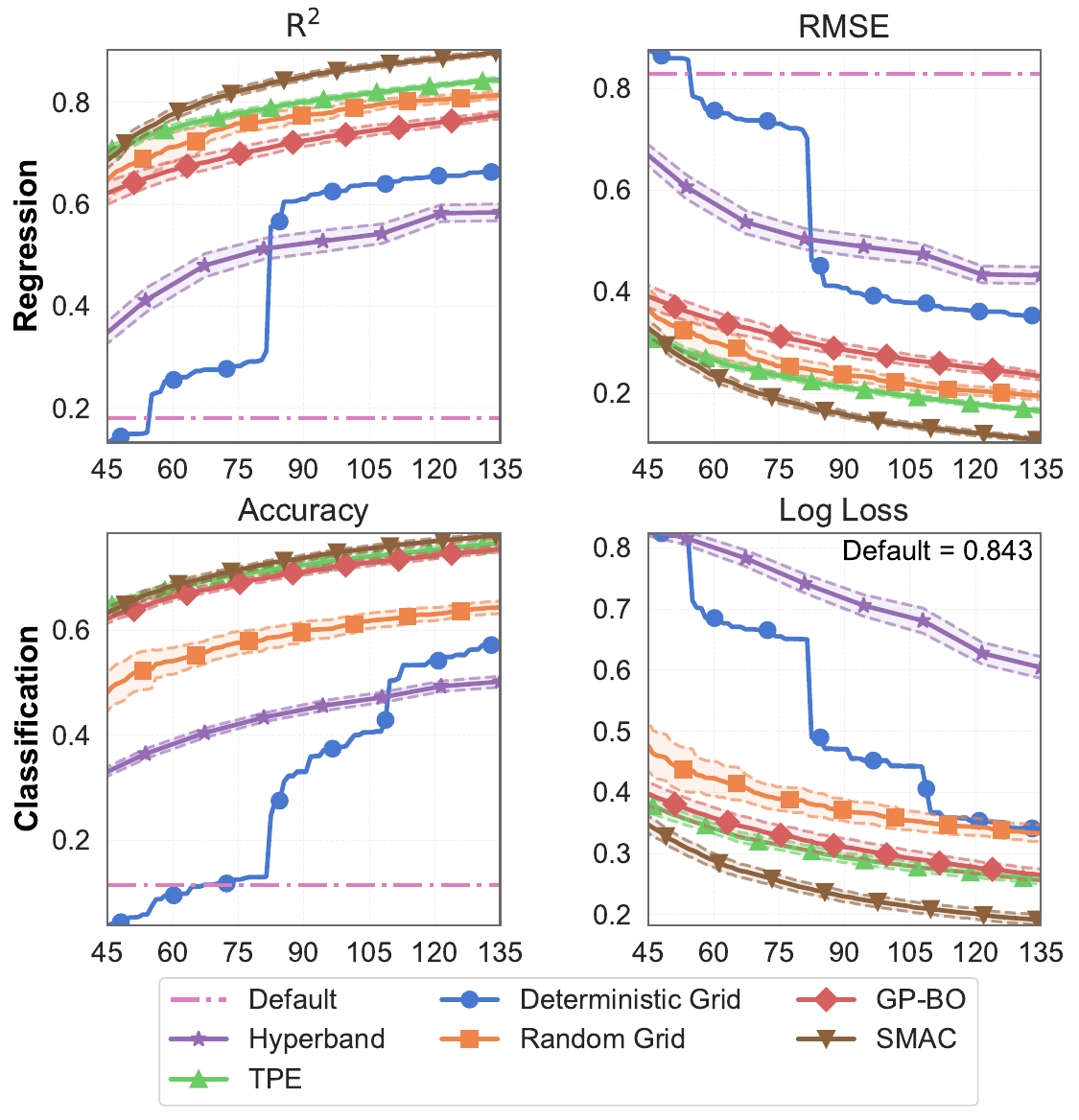}
    \caption{Normalized scores as a function of the number of trials. The confidence intervals represent uncertainty due to the randomness in the hyperparameter selection methods.}
    \label{fig:comparison_num_leaves_seed}
\end{figure}

\begin{figure}[ht!]
    \centering
    \includegraphics[width=\linewidth]{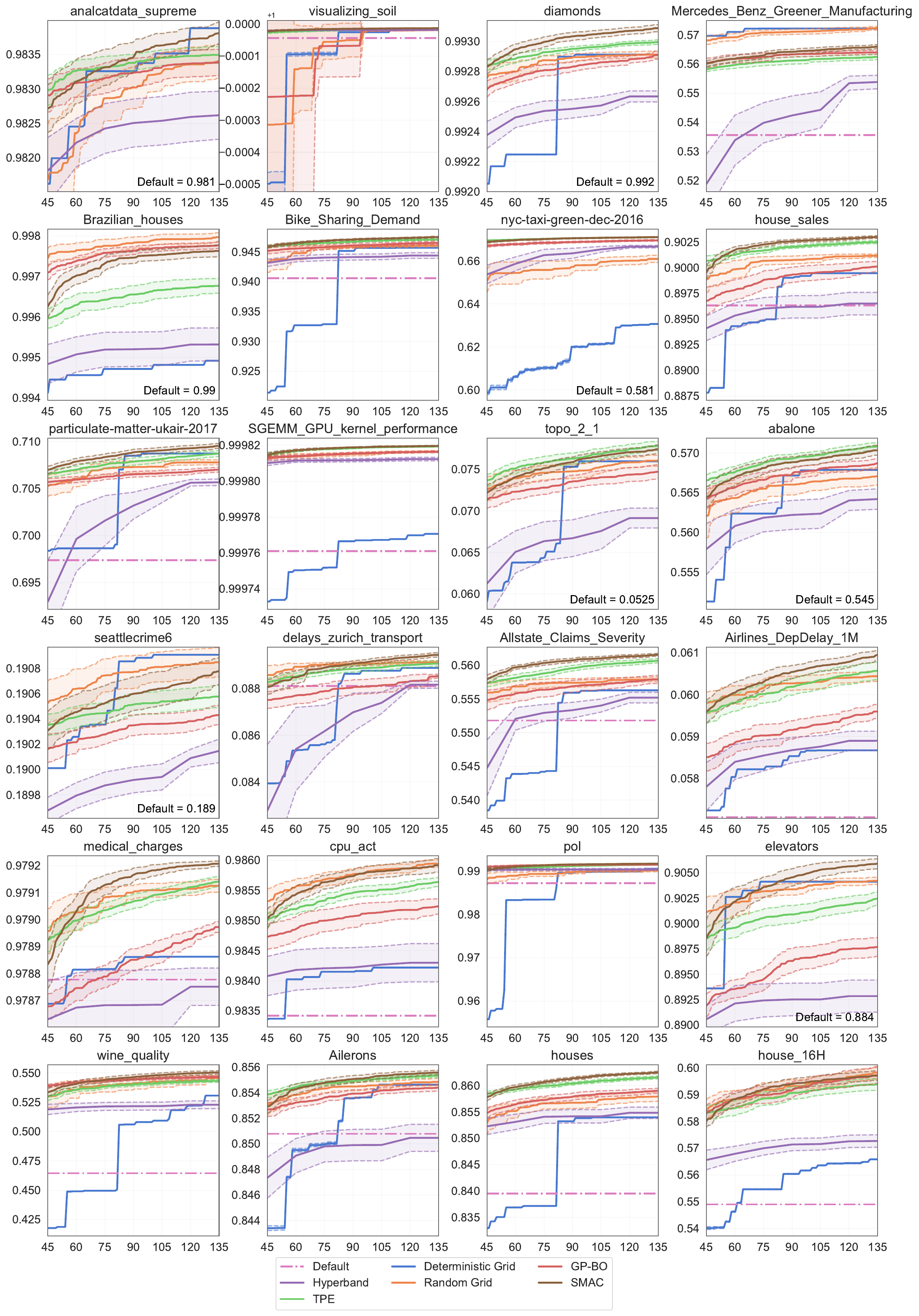}
    \caption{$R^2$ as a function of the number of trials per data set. The confidence intervals represent uncertainty due to the randomness in the hyperparameter selection methods.}
    \label{fig:comparison_by_task_num_leaves_R2_1}
\end{figure}

\begin{figure}[ht!]
    \centering
    \includegraphics[width=\linewidth]{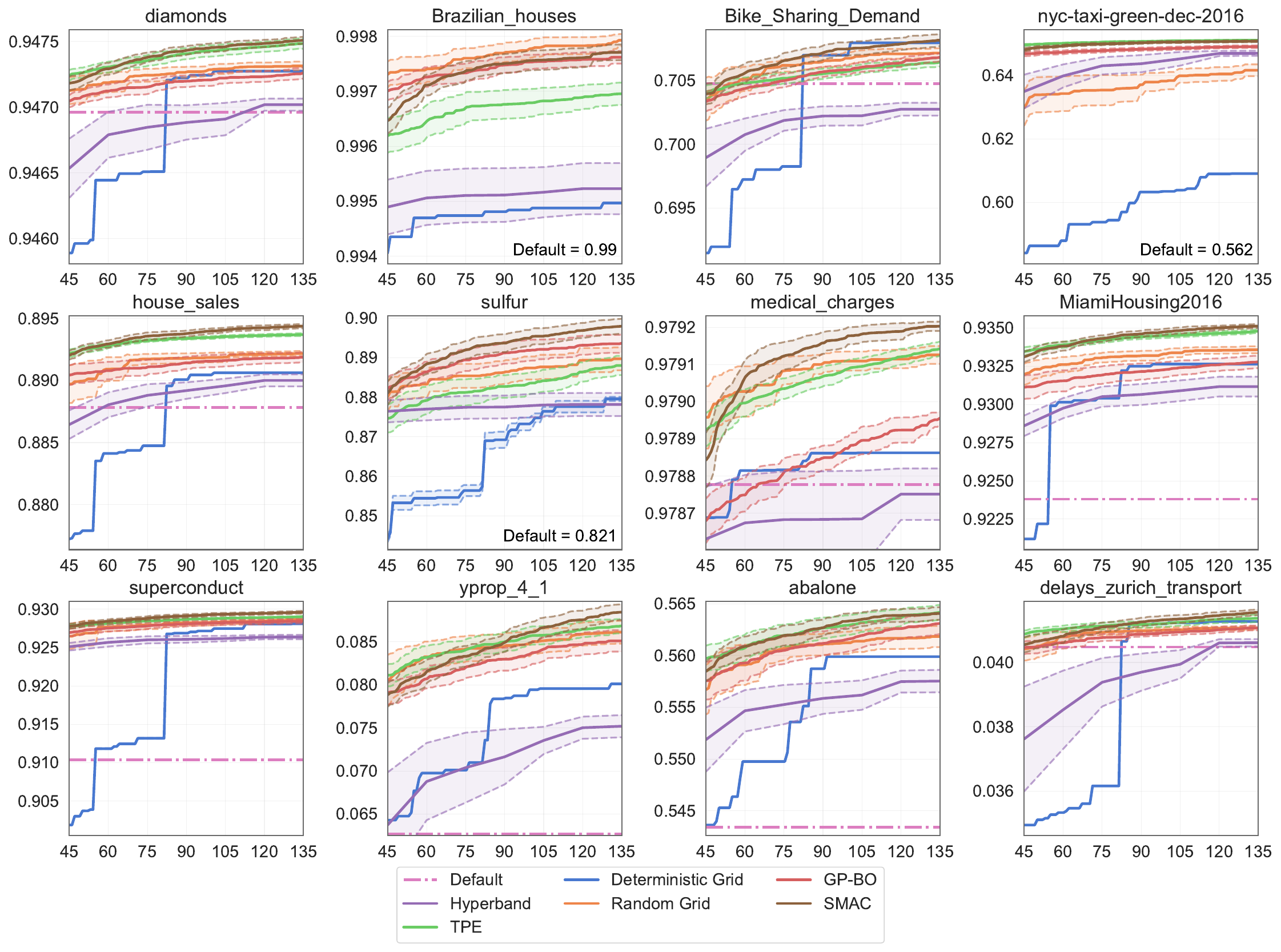}
    \caption{$R^2$ as a function of the number of trials per data set. The confidence intervals represent uncertainty due to the randomness in the hyperparameter selection methods.}
    \label{fig:comparison_by_task_num_leaves_R2_2}
\end{figure}

\begin{figure}[ht!]
    \centering
    \includegraphics[width=\linewidth]{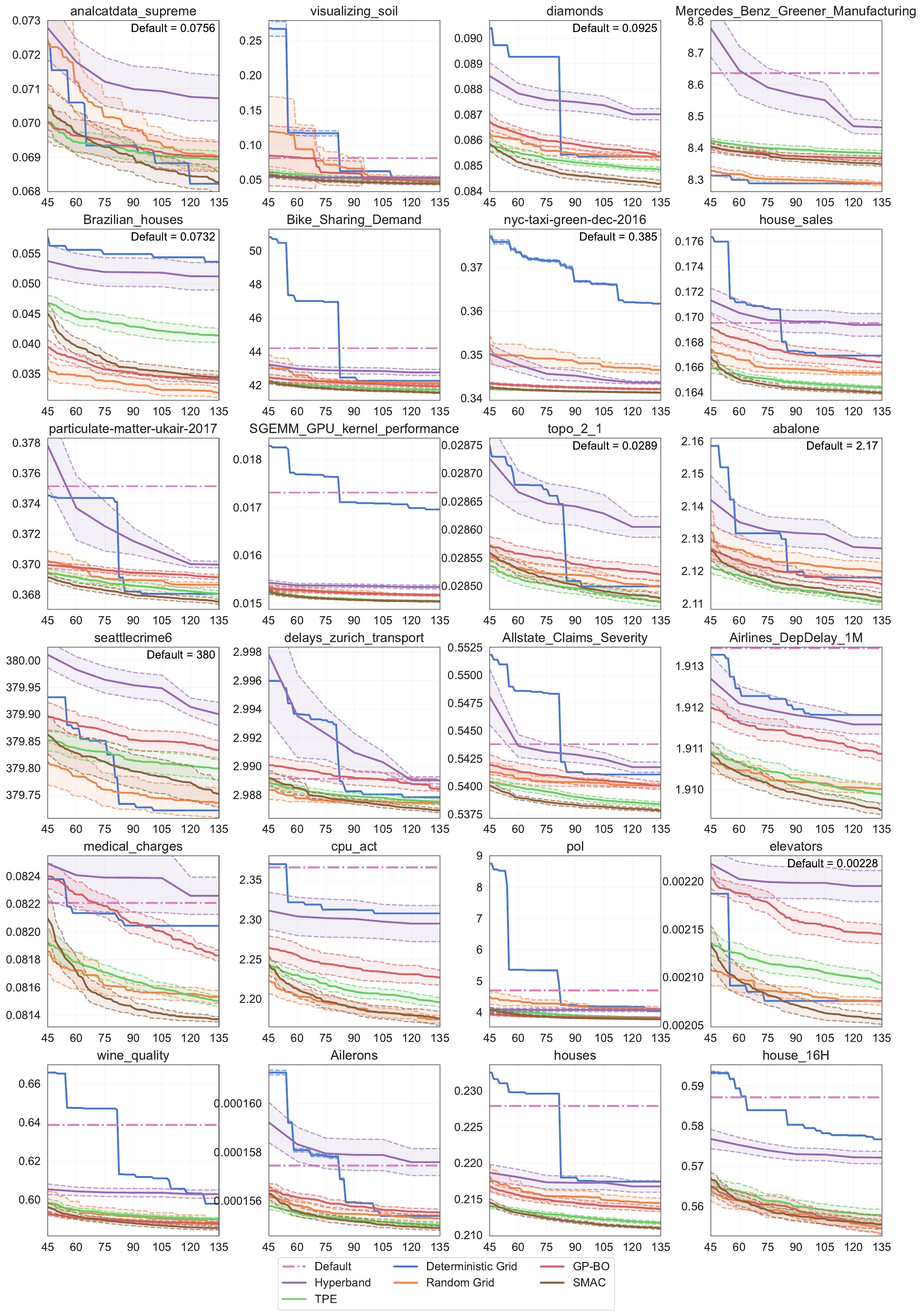}
    \caption{$\mathrm{RMSE}$ as a function of the number of trials per data set. The confidence intervals represent uncertainty due to the randomness in the hyperparameter selection methods.}
    \label{fig:comparison_by_task_num_leaves_RMSE_1}
\end{figure}

\begin{figure}[ht!]
    \centering
    \includegraphics[width=\linewidth]{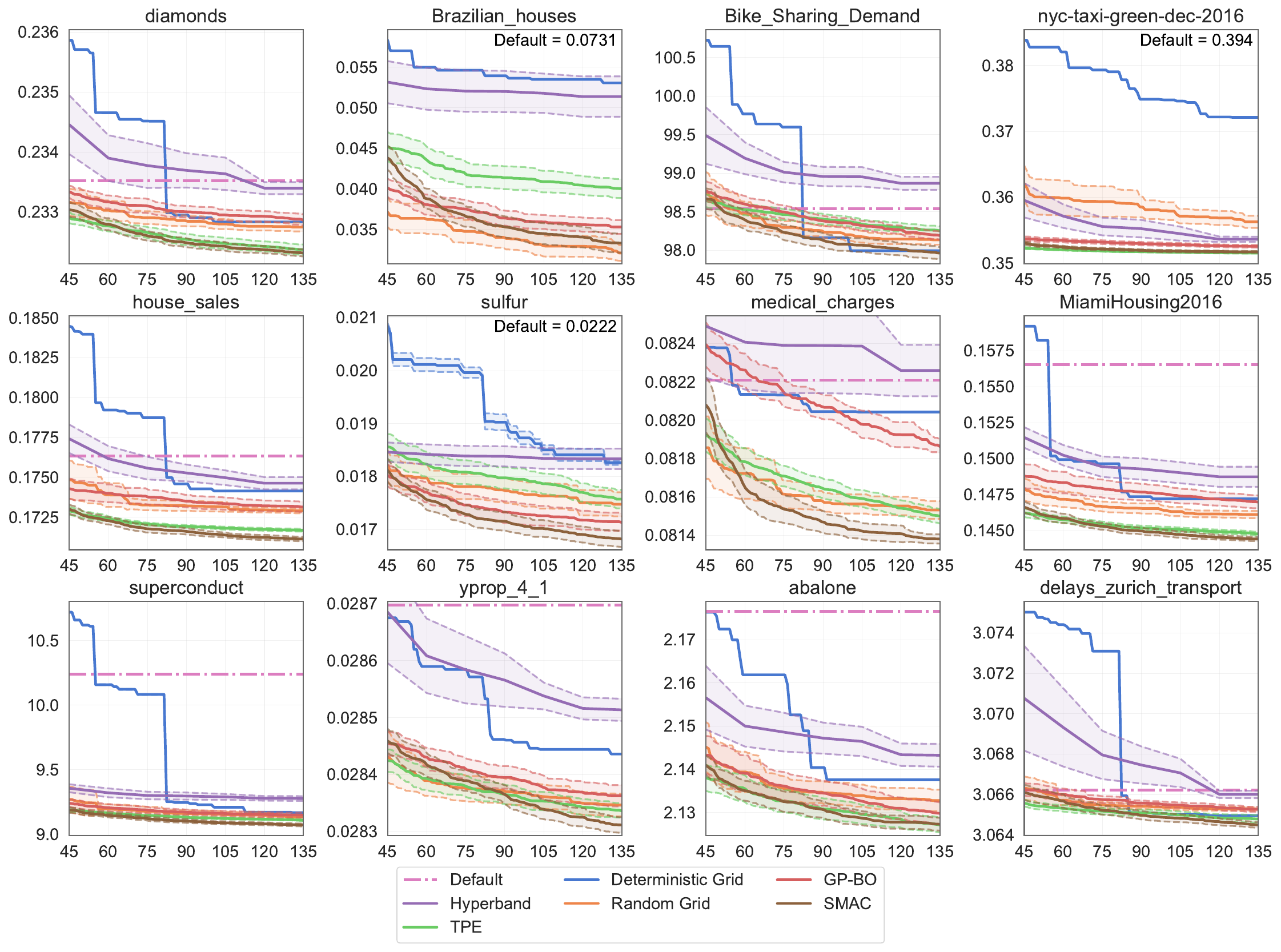}
    \caption{$\mathrm{RMSE}$ as a function of the number of trials per data set. The confidence intervals represent uncertainty due to the randomness in the hyperparameter selection methods.}
    \label{fig:comparison_by_task_num_leaves_RMSE_2}
\end{figure}

\begin{figure}[ht!]
    \centering
    \includegraphics[width=\linewidth]{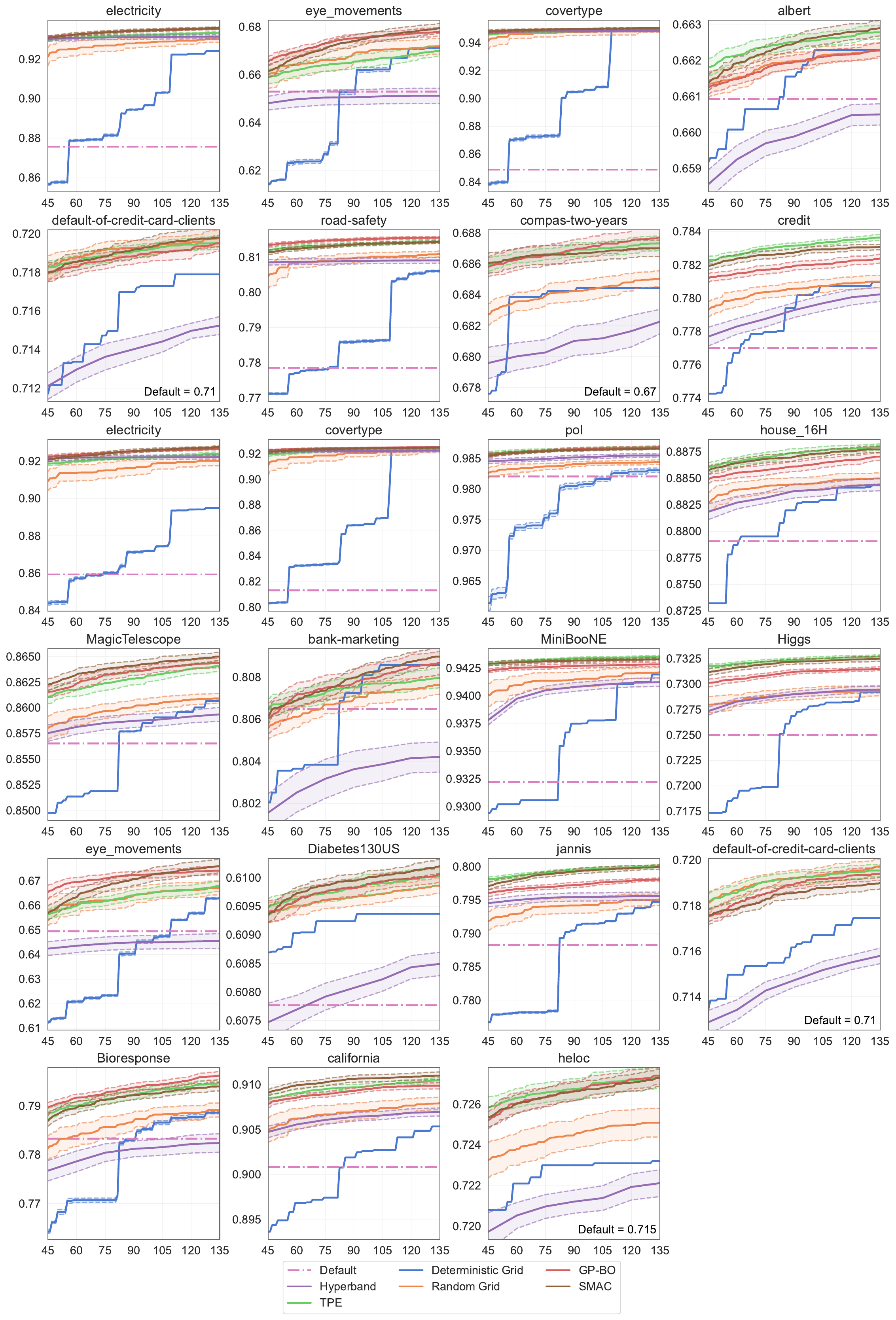}
    \caption{Accuracy as a function of the number of trials per data set. The confidence intervals represent uncertainty due to the randomness in the hyperparameter selection methods.}
    \label{fig:comparison_by_task_num_leaves_acc}
\end{figure}

\begin{figure}[ht!]
    \centering
    \includegraphics[width=\linewidth]{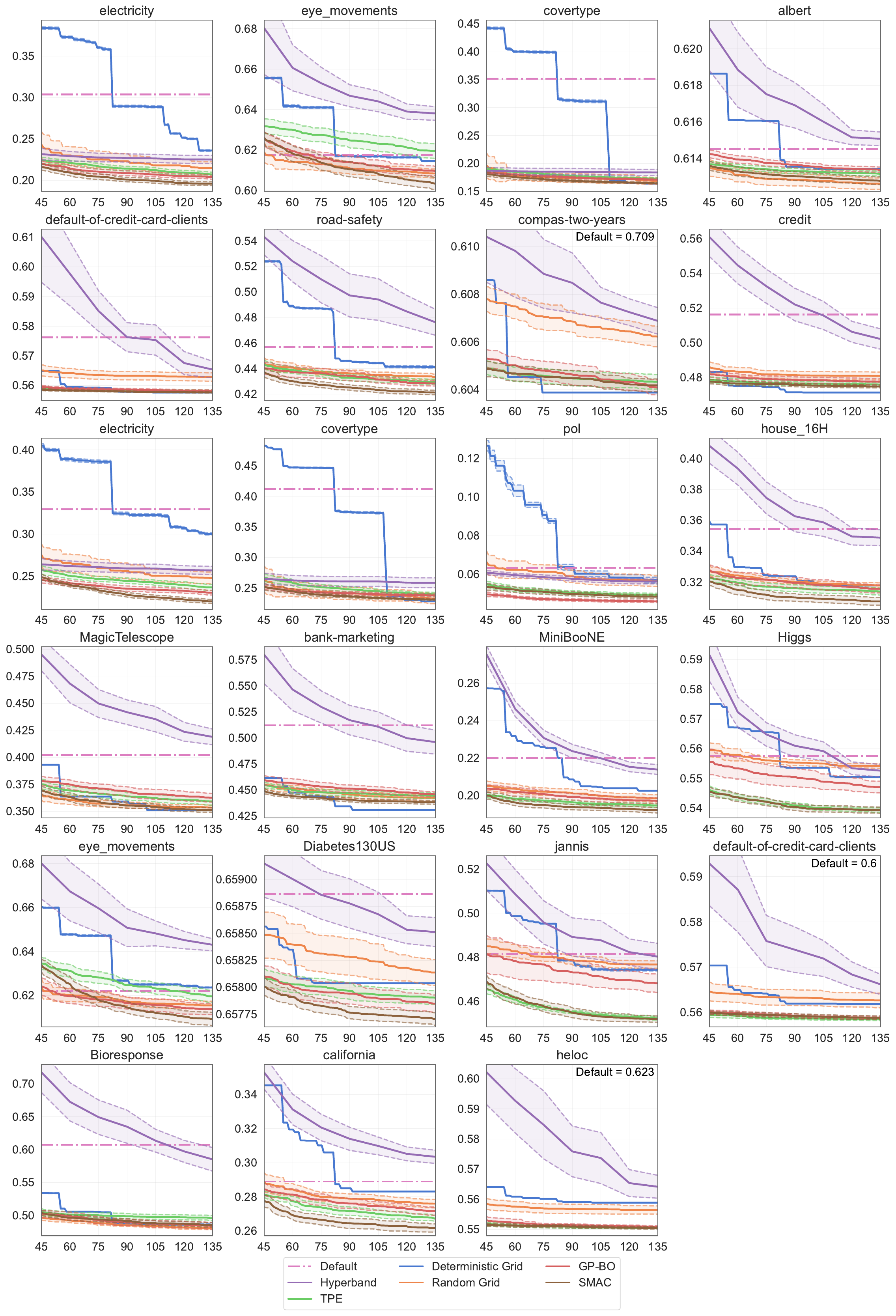}
    \caption{Log Loss as a function of the number of trials per data set. The confidence intervals represent uncertainty due to the randomness in the hyperparameter selection methods.}
    \label{fig:comparison_by_task_num_leaves_log_loss}
\end{figure}


\begin{figure}[htbp]
    \centering
        \includegraphics[width=\linewidth]{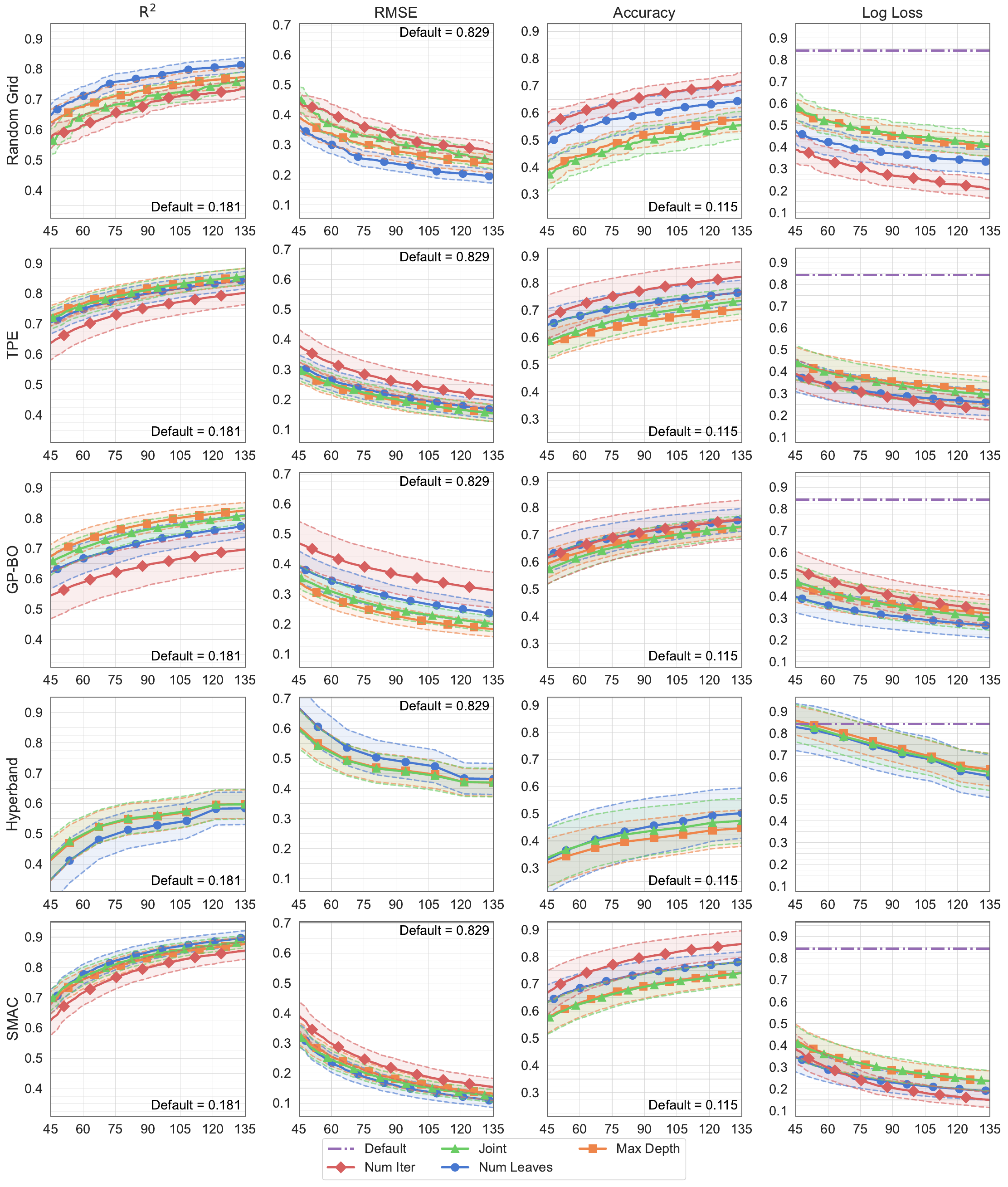}
    \caption{Comparison of hyperparameter search space options regarding the maximal number of iterations, the maximal tree depth, and the number of boosting iterations.}
    \label{fig:comparison_tuning_grid_regr}
\end{figure}


\end{document}